\documentclass[fleqn,10pt]{wlpeerj}

\title{Urdu text in natural scene images: A new dataset and preliminary text detection}

\author[1]{Hazrat Ali}
\author[2]{Khalid Iqbal}
\author[1]{Ghulam Mujtaba}
\author[1]{Ahmad Fayyaz}
\author[3]{Mohammad Farhad Bulbul}
\author[1]{Fazal Wahab Karam}
\author[1]{Ali Zahir}

\affil[1]{Department of Electrical and Computer Engineering, COMSATS University Islamabad, Abbottabad Campus, Abbottabad, Pakistan.}
\affil[2]{Department of Computer Science, COMSATS University Islamabad, Attock Campus, Attock, Pakistan.}
\affil[3]{Department of Mathematics, Jashore University of Science and Technology, Jashore-7408, Bangladesh}
\corrauthor[1]{Hazrat Ali}{hazratali@cuiatd.edu.pk}

\begin{abstract}
Text detection in natural scene images for content analysis is an interesting task. The research community has seen some great developments for English/Mandarin text detection. However, Urdu text extraction in natural scene images is a task not well addressed. In this work, firstly, a new dataset is introduced for Urdu text in natural scene images. The dataset comprises of 500 standalone images acquired from real scenes. Secondly, the channel enhanced Maximally Stable Extremal Region (MSER) method is applied to extract Urdu text regions as candidates in an image. Two-stage filtering mechanism is applied to eliminate non-candidate regions. In the first stage, text and noise are classified based on their geometric properties. In the second stage, a support vector machine classifier is trained to discard non-text candidate regions. After this, text candidate regions are linked using centroid-based vertical and horizontal distances. Text lines are further analyzed by a different classifier based on HoG features to remove non-text regions. Extensive experimentation is performed on the locally developed dataset to evaluate the performance. The experimental results show good performance on test set images. The dataset is available for research use. To the best of our knowledge, the work is the first of its kind for the Urdu language and would provide a good dataset for free research use and serve as a baseline performance on the task of Urdu text extraction. 
\end{abstract}

\begin{document}

\flushbottom
\maketitle
\thispagestyle{empty}

\begin{tabular}{p{12cm}}
\hline
\textbf{Cite as}: Ali H, Iqbal K, Mujtaba G, Fayyaz A, Bulbul MF, Karam FW, Zahir A. 2021. Urdu text in natural scene images: a new dataset and preliminary text detection. PeerJ Computer Science 7:e717 https://doi.org/10.7717/peerj-cs.717 \\
\hline
\end{tabular}

\section*{Introduction}

Recent developments within the computer vision and machine learning research communities have witnessed advancement in the task of text detection within natural scene images. Detection of text in natural scene images has applications ranging from robot navigation to understanding of sign prompts by self-driving cars \citep{yuan2016incremental}. Text detection in natural scene images is a challenging task due to background complexity, occlusion, different font styles and orientations, variations in scale, and variations in illumination. Detection of text in such complex scenarios leads to false positives \citep{yao2012detecting}. Most of the work regarding text detection in scene images has been for English and Chinese languages facilitated by the availability of numerous benchmark datasets \citep{zhu2016scene}. However, the detection of cursive text (such as Urdu text) from natural scene images has not been investigated yet due to the unavailability of suitable datasets for the training and testing of models. 

Some notable developments on text detection can be found in  \citep{iqbal2014bayesian}, \citep{matas2004robust}, \citep{shi2013scene}, \citep{he2016text}, \citep{iqbal2017ztext}, \citep{yin2012effective}, \citep{iqbal2013classifier}, \citep{shahab2011icdar}.
Text detection methods, in general, can be classified into two broad categories: window-based techniques and connected component-based techniques. In window-based techniques, low-level features are extracted by scanning the image several times at different scales. These algorithms, i n general, are computationally demanding and time-consuming \citep{zhang2013text}. The connected component based algorithms have been categorized into three classes: region-based methods, texture-based methods, and stroke width transform (SWT) methods \citep{zhang2013text}. In region-based methods, edges and textual properties of text are reliable features that are used in text extraction. Texture-based methods use feature-based approaches to calculate gray level co-occurrence matrix (i.e., calculate features such as homogeneity, dissimilarity, and contrast). In stroke width transform methods, the per-pixel width is considered as a local image operator \citep{epshtein2010detecting}. Stroke is considered as the contiguous part of an image with constant width.  Typically, stroke is extracted through image segmentation, extraction of features, and classification. Following this, individual strokes are grouped through clustering. Once a text is detected, text regions are identified by drawing bounding boxes. The coordinates of the bounding boxes can help to locate the text region for extraction. Several approaches for text detection and extraction are discussed in \citep{zhang2013text}. Segmentation is used for extraction of the region of interest despite the fact of having background complexity in natural scene images. Thresholding, color clustering, and statistical-based methods have typically been popular, as discussed in \citep{zhang2013text}. 

\textbf{Urdu language:} There has been a substantial amount of research work reported on text processing of major languages, but no significant work has done been so far for Urdu text detection. Urdu is the national language of Pakistan. It is spoken in more than 20 nations of the world. The number of speakers of Urdu is in the range of 60 to 70 million. The writing style of Urdu is from right to left. Arabic and Farsi languages have some similarities in writing styles, however, Urdu poses more challenges as it possesses more characters than Arabic and some of these characters have complicated writing styles due to additional diacritics or differences in shapes \citep{khan2012efficient}. The distinctive characteristics of the Urdu language are; (a) There are no uppercase or lowercase letters. (b) Unlike English, Urdu is written from right to left. (c) Unlike the English text, the words of Urdu are not squared in shape 

One key reason behind the rare consideration of Urdu text retrieval is the unavailability of the Urdu text images dataset. Therefore, one of our contributions is a benchmark dataset for researchers to kick start research development on Urdu text extraction and processing. Besides, our proposed framework can be a baseline to compare Urdu text detection in natural scene images. Hence, we propose a framework for Urdu text detection and extraction in natural scene images contained in a dataset developed locally from real scene images. We use channel enhanced maximally stable extremal region (MSER) for text region detection. A two-stage filtering mechanism is applied to eliminate candidate regions. In the first stage, text and noise are classified based on their geometric properties. In the second stage, an SVM classifier is trained to discard non-text candidate regions. After this, text candidate regions are linked using centroid-based vertical and horizontal distances. Text lines are further analyzed by a different classifier based on HOG features to remove non-text regions. For classification, an SVM model is trained with the features obtained from the text regions. 

The main contributions of this work are:
\begin{itemize}
    \item This work presents a new dataset of natural scene images having Urdu text in different fonts and styles.
    \item The dataset comprises 500 natural images that are available freely for research use.
    \item This work highlights the challenges for Urdu text detection in natural scene images and identifies the knowledge gap due to which research progress on the topic has been slow.
    \item This work presents a baseline framework on MSER features for the extraction of Urdu text in natural scene images.
\end{itemize}


\section*{Related Work}
\label{sec:relatedwork}
Text detection and recognition have greatly been developed for English-based scanned documents, books, and natural scene images \citep{plamondon2000online}. Compared to many other languages, English text detection is relatively less challenging due to the simple style of English letters combination. In \citep{jain2014text}, English text is extracted together through the maximally stable extremal region (MSER) features and the use of the Markov Model approach along with support vector machines (SVM). Other approaches used for extraction of English text within scene images include multi-scale thresholding, SWT approaches, conditional random field model approaches, and hybrid approaches with improved recognition accuracy, as discussed in \citep{jain2014text}. Besides English, considerable work has been carried out on Chinese text extraction. For example, Chinese text extraction has been performed using MSER-based techniques along with clustering approaches \citep{jain2014text}.
While there is no significant work on Urdu text extraction from natural scene images, it is still useful to briefly summarize the work done for many similar languages such as Arabic, Persian and Uyghur languages. Ahmed et al., \citep{ahmed2017deep} have reported a text recognition approach using convolutional neural networks for Arabic text in scene images. The work mainly addressed the recognition of isolated characters of Arabic text within scene images after a total of 2700 characters have been extracted by segmentation for 27 unique characters (i.e., 100 images per character). Yan et al., \citep{yan2017effective} proposed a dataset named IMAGE570 containing Uyghur text scene images. The dataset IMAGE570 contained captioned text images along with scene text images in which 370 images are kept for training data and 200 images for testing data. For text and non-text regions proposal, they proposed channel enhanced MSER technique and for classification, they trained SVM with HOG features.
In \citep{darab2012hybrid}, a hybrid approach is applied to Persian text detection in scene images. A large dataset of Persian text is presented and state-of-the-art results have been reported. Edge-based features and color information of regions in images are computed for regions proposal. HOG features along with wavelet histograms are computed for text region classification. A local features descriptor technique is applied to detect and recognize text of Korean and Japanese languages. The approach has used scale-invariant feature transform (SIFT) features for extracting text \citep{zheng2010text}.
Jamil et al., \citep{jamil2011edge} proposed edge-based features for Urdu text localization in captioned text video images. The proposed approach for Urdu text detection in captioned text videos is based on edge-based segmentation and few heuristic rules. The detected text regions are then segmented from the background. A dataset of Urdu captioned text video images is proposed by Imran et al., containing 1000 images of videos \citep{raza2012database}. There has been an attempt reported for Urdu handwritten characters recognition by \citep{ali2020pioneer}. However, this work addresses individual characters recognition on clean background and text detection in natural scene images is not addressed. Recently, dataset for Urdu text in scene images has been proposed for character extractions and recognition \citep{chandio2018character}. The dataset has 600 Urdu text scene images. Text region is manually cropped from these images and made into a dataset of 18000 Urdu characters of 48×48 dimensions for training and testing purpose. Different classifiers are trained with HOG features and efficiency of these classifiers on the proposed dataset is evaluated. Another relevant work on Urdu text recognition is reported by \citep{arafat2020IEEE}. The authors have used a deep residual network (ResNet18) for recognition of Urdu text. However, their main contribution is for recognition of text in outdoor images with synthetic text. The synthetic text is a computer-generated text with uniform fonts and size. The results for text recognition in natural scene images are limited to 76.6\% accuracy rates on a test set of 110 images.

From the literature review of text detection in natural scene images, we found that little work has been done so far for Urdu text detection in natural scene images. We also found that there is no public dataset available for Urdu text detection in natural scene images. To address these issues and fill the research gap, in this work, we propose a new dataset for Urdu text detection, which is available to the research community for further research purposes. We also contribute a framework for Urdu text detection in scene images with state-of-the-art results.

\section*{Dataset}
\begin{figure}[ht!]
\centering
\includegraphics[width=0.7\linewidth]{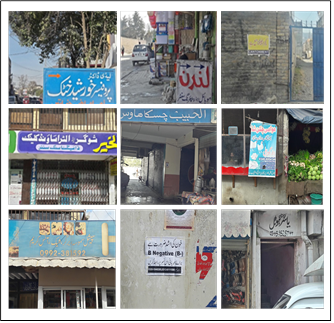}
\caption{Examples of images from the dataset}
\label{fig:sampleimages}
\end{figure}

\begin{figure}[ht!]
\centering
\includegraphics[width=0.7\linewidth]{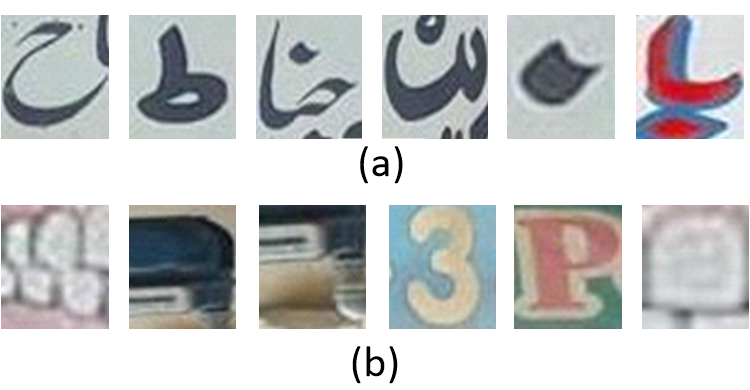}
\caption{(a) Urdu text regions and (b) Urdu non-text regions}
\label{fig:image2}
\end{figure}

We captured 500 Urdu text scene images in 1080 × 800 resolution. The images are acquired through a smartphone camera with no camera built-in pre-processing filters. The images are acquired in different times of the day to cover diverse lightning conditions. This helps in developing a dataset with images recorded under realistic conditions. Some of the images of the dataset are shown in Figure \ref{fig:sampleimages}. Out of these, 400 images are used for training and the remaining 100 images are used for testing purpose. From the training dataset, we manually cropped 7500 patches of text-regions of size 42 × 46 and 14500 non-text regions for training the classifier. Samples of the cropped text and non-text images are shown in Figure \ref{fig:image2}.

\begin{figure}[ht!]
\centering
\includegraphics[width=0.7\linewidth]{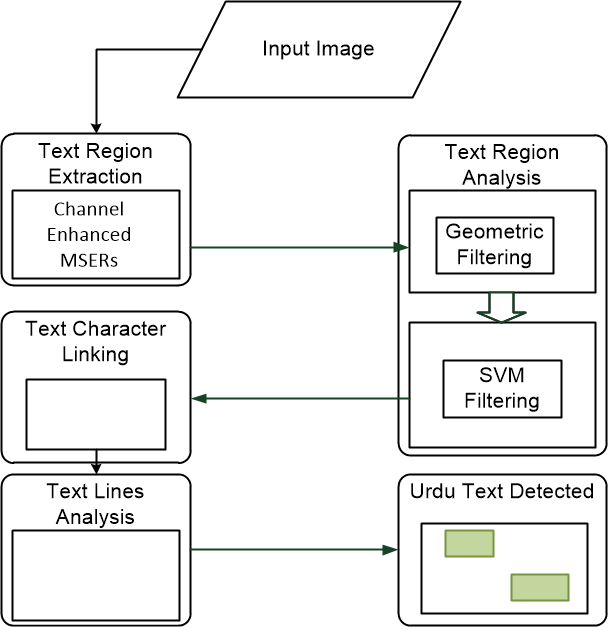}
\caption{The workflow of the proposed model. The four modules are: text region extraction, text region filtering, text character linking, and text lines analysis.}
\label{fig:workflow}
\end{figure}

\begin{figure}[ht!]
\centering
\includegraphics[width=0.8\linewidth]{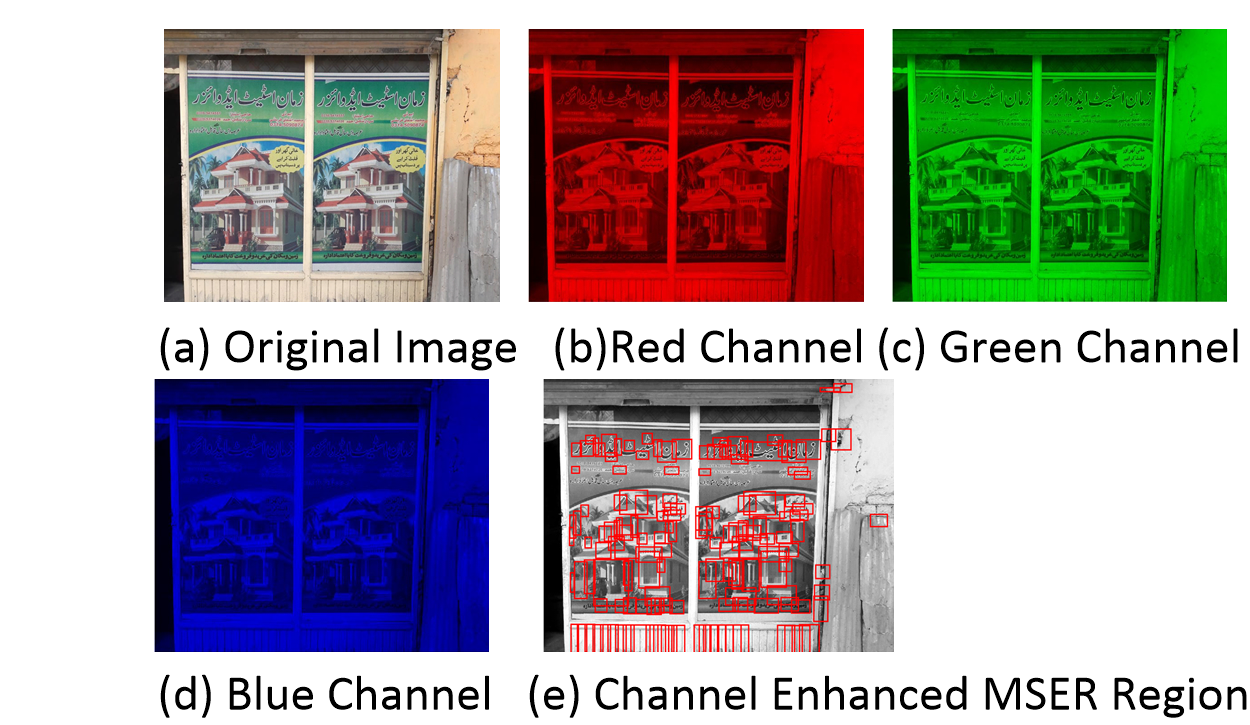}
\caption{Regions detected through channel enhanced MSER. In the channel enhanced method, the MSER regions are detected separately for text in each color channel and then the detection regions are combined. The detected regions are shown through red boundary boxes.}
\label{fig:channels}
\end{figure}

\begin{figure}[ht!]
\centering
\includegraphics[width=0.7\linewidth]{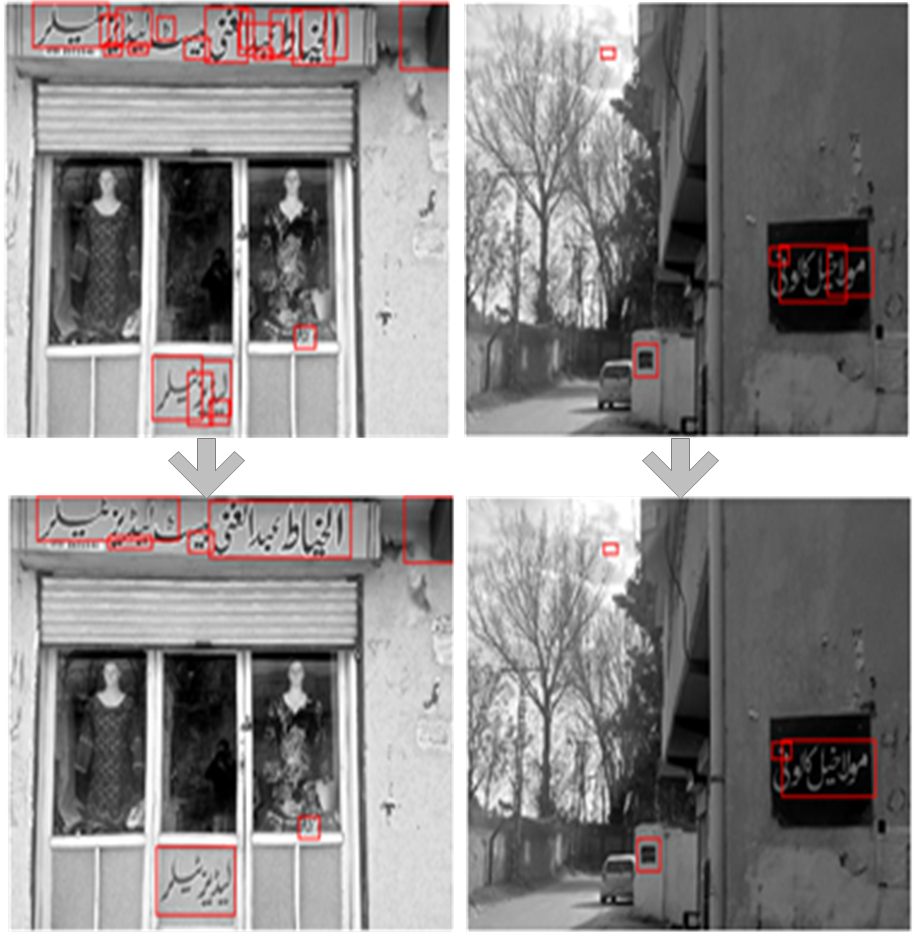}
\caption{Text regions after SVM classification (Top row). Candidate text regions fulfilling the criteria for text character linking are merged into text lines.}
\label{fig:regionmerging}
\end{figure}

\begin{figure}[ht!]
\centering
\includegraphics[width=0.7\linewidth]{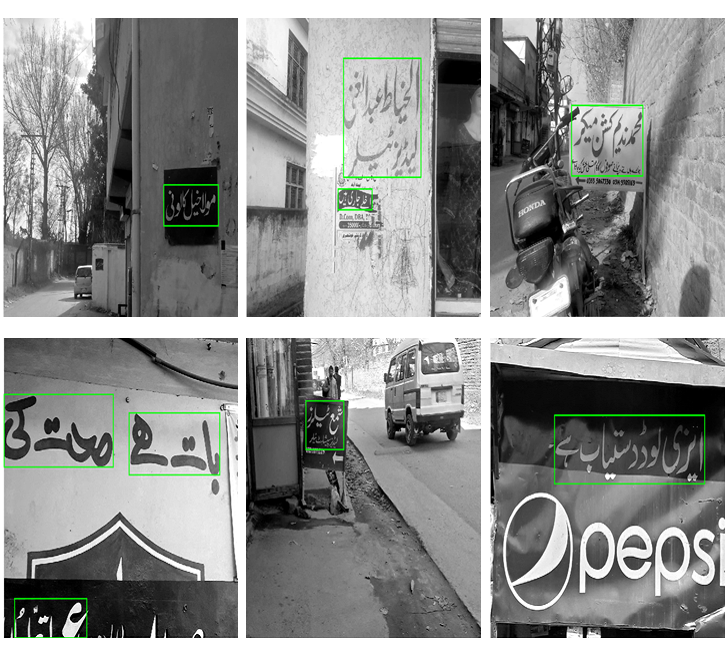}
\caption{Sample images showing ground truth Urdu text regions}
\label{fig:groundtruth}
\end{figure}

\section*{Methodology}
The overall workflow for the proposed model has four main working modules, as shown in Figure \ref{fig:workflow}. These four modules are; text region extraction, text region filtering, text character linking, text lines analysis. We discuss each module below. 

\subsection*{Text Region Extraction}
The connected component-based approaches are popular for their low computational cost. On the other hand, frequency-based approaches for text region extraction are time demanding and computationally expensive. We use the channel enhanced MSER method for Urdu text regions extraction. The channel enhanced MSER technique is popular for its robustness and efficiency. It is adopted for text regions localization. MSER is a connected part of an image whose pixels have a consistent higher or lower intensity than its outer boundary pixels. 
The MSER extracted region captures the features such as stroke width, aspect ratio, eccentricity, solidity, extent, and Euler number. MSER typically captures the image regions where the local contrast is high.

The MSER works well on grayscale images. However, for blurry images, the performance is poor due to a lack of contrast between the background and the foreground. This is common when RGB images are converted to grayscale images. But in general, most of the foreground has high contrast while the background contents mainly occur in one of the RGB channels than in grayscale. So, Yan et al. proposed channel enhanced MSER region for text detection \citep{yan2017effective}. In the channel enhanced method, the MSER regions are detected separately for text in each color channel, and then the detection regions are combined. In this work, for the text regions' proposal, we apply the channel enhanced MSER method. Figure \ref{fig:channels} shows the channel enhanced MSER result on a sample image.

\subsection*{Text Region Filtering}
Regions generated by channel enhanced MSER techniques detect true positives along with a large number of false positives. To remove false positive and retain true positives only, a two-step filtering technique is applied. 

\subsubsection*{Geometric Filtering}
After detecting MSER regions, we apply a few heuristic rules to ﬁlter out the non-text regions. Different thresholds are used to discard noise. Thresholds values are not fixed but instead selected manually for a dataset. In geometric filtering, regions are discarded based on geometric properties such as stroke width variation, aspect ratio, eccentricity, solidity, extent, and Euler number \citep{brooks2017exploring}. The threshold values of these features are mostly data specific. About 20 to 25 percent of non-text regions are discarded at this stage. The heuristic rules for geometric filtering adopted in this work are discussed below.

\textbf{Overlapping ratio:} If two regions share a common region with an overlap ratio greater than 80 percent, then the region with more significant variations is discarded. This helps in filtering out repetitive regions.  

\textbf{Large regions:} As we know from MSER, a larger candidate region may have few but not too many smaller candidate regions. So, we define a threshold value to check if the number of small candidates is greater than the specified threshold value. If so, the algorithm discards certain small regions considering them as non-text regions.

\textbf{Aspect ratio:} The ratio between width and height is called the aspect ratio. It is important in discarding non-text regions. It discards regions whose aspect ratio is different from text because the text has a well-defined aspect ratio. In contrast, non-text regions in the background have lower or higher aspect ratios typically.

\textbf{Eccentricity:} The distance between the major axis of the ellipse and the center is called eccentricity. It is helpful for the measurement of the circular properties of the candidate regions. It is also widely used for non-text regions filtering. In text detection, it provides robustness against noise. It is helpful for text region detection as the value provides unique characteristics of the text region.

\textbf{Euler number:} The number of holes subtracted from the number of connected objects is known as the Euler number. In the case of only one region, it is subtracted from one. As the text candidate region can have only a certain range for the Euler number value, it is useful to discard those regions whose number of holes is greater than text candidate regions.

\textbf{Stroke width variation:} In general, the stroke width is smooth for text regions and highly variant for the background. Hence, it is a useful parameter for the elimination of non-text regions. The approach is to set a threshold value of the variance of stroke width. The regions with stroke width variance greater than the threshold value are regarded as noise and hence discarded. It is a widely used criterion for filtering non-text regions in scene images. Still, it cannot be used for Urdu text extraction because the stroke width variation is also high in Urdu text candidates.

\subsubsection*{SVM Filtering}
Channel enhanced MSER algorithms detected the text and non-text regions. In geometric filtering, about 20 to 30 percent of non-text regions are discarded successfully, but further filtering is required to discard more non-text regions. In this filtering stage, we use an SVM classifier that discards those non-text regions, left after the first stage. SVM is trained with unsupervised features extracted through k-means clustering \citep{pan2010fast}. SVM classifier is trained with 23000 training images of dimension 42×46×3 having 7500 text candidates and 14500 non-text candidates (performance obtained with other dimensions are discussed in the results section). At this stage, most of the non-text regions are discarded but few false-positive candidates are left.

\subsection*{Text Characters Linking}
After geometric and SVM classification, we have mostly all true positive candidates left along with a few false positives. At this stage, we link text characters into pairs and then into lines based on the following rules: 

\begin{enumerate}[noitemsep] 
\item First, we compute the centroids of each region.
\item The Y difference of two centroids must not be greater than the height of the smaller character for merging them into pair.
\item Also, the X difference between two centroids must not be greater than 2 times the height of the larger region.
\end{enumerate}
These conditions ensure that only horizontal and closely located text regions become part of text lines. Regions fulfilling the criteria are combined into text lines, as shown in Figure \ref{fig:regionmerging}.

\subsection*{Text Lines Analysis}
After first layer filtering that includes geometric filtering followed by SVM filtering, we got most of the text regions and a few also few non-text candidate regions. Then, in the text character linking stage, we combined text candidate regions to detect horizontal text lines. At this point, we achieve a high recall rate but low precision value. It means that we have detected text lines with more accuracy, but we also have non-text regions detected in the background. So, in the filtering layer of text lines analysis, the detected text lines are verified through different classifiers trained with the histogram of oriented gradients (HOG) features. 

\section*{Results}
The performance of the framework is evaluated on the test set comprising of 100 images containing Urdu text in complex background. The performance evaluation technique is the one proposed in Yan et al., \cite{yan2017effective}. It is critical to evaluate whether a text region is identified correctly. For the detected text rectangle (DT) and the ground truth rectangle (GT), the overlap ratio proportion is calculated to determine whether a text region is correctly identified. The overlap ratio between DT and GT is deﬁned as:
\begin{align}
\centering
    Overlap \; Ratio = \frac{Area(GT \cap DT)}{Area(GT \cup DT)}
\end{align}
If the overlap ratio between ground truth and detected text line is greater than 0.5, then it is considered as true positive; otherwise, the detected text line is a false positive. For single-line text, if we have more non-repeating detected text lines, then DT is the union of all these detected text lines. In the case of multiline text, then GT is the union of all text lines. Once the numbers of true and false positives are computed, we then we measure the performance in terms of precision (p), recall (r), and F-measure (f), as shown in the Equation \ref{eq:precision} through Equation \ref{eq:F} and also used by other works on detection and classification \citep{do2021using, le2020xgboost}.

\begin{align}
\centering
Precision,\; p = \frac{\mid TP \mid}{\mid E \mid}
\label{eq:precision}
\end{align}

\begin{align}
Recall,\; r = \frac{\mid TP \mid}{\mid T \mid}
\label{eq:recall}
\end{align}

\begin{align}
F\; score = \frac{2pr}{(p + r)}
\label{eq:F}
\end{align}

Here $TP$ represents true positive detected lines, $E$ represents estimated rectangles and $T$ represents ground truth rectangles. 
Sample images of ground truth are shown in Figure \ref{fig:groundtruth}. We have manually cropped 7500 text characters and 14000 non-text candidates from 400 images of the training dataset for feature extraction and training classifiers. To compare performances, the detected regions with channel enhanced MSER are resized to three different dimensions i.e., $40\times36$, $46\times42$, and $54\times50$. 

For evaluation purposes, the manually cropped text and non-text regions from training images are divided into training and test subsets. 5000 images of both text and non-text regions are selected for training and 1000 images for each class are kept for testing. For final classification, we train different SVM models. Table \ref{tab:tab1}, Table \ref{tab:tab2}, and Table \ref{tab:tab3} show the experimental results for SVM with linear, Gaussian (radial basis function), and polynomial degree 3 kernels, respectively. The three different resized dimensions are $40\times36$, $46\times42$, $54\times50$. From the comparison, we find that SVM with polynomial kernel on 40×36 resized training images achieve the highest accuracy of \textbf{0.8880}. 

\begin{table}[ht!]
\centering
\begin{tabular}{l|l|l|l|l}
Dimension & Precision & Recall & F measure & Accuracy \\\hline
$40\times36$ & 0.8479 &	0.8454	& 0.8467 & 0.8462 \\
$46\times42$ &	0.8424 & 0.8424	& 0.8424 & 0.8417 \\
$54\times50$ &	0.8498	& 0.8414 &	0.8456 & 0.8457 \\ \hline
\end{tabular}
\caption{\label{tab:tab1}Performance of SVM with Linear Kernel}
\end{table}
\begin{table}[ht!]
\centering
\begin{tabular}{l|l|l|l|l}
Dimension &     Precision &     Recall &    F measure &     Accuracy \\\hline
$40\times36$ &  0.8889 &	    0.8801	&   0.8845	&   0.8845 \\
$46\times42$ &	0.8818 &	    0.8722 &	0.8769 &	0.8771 \\
$54\times50$ &	0.8909	&       0.8741	&   0.8824	&   0.8830 \\ \hline
\end{tabular}
\caption{\label{tab:tab2}Performance of SVM with Gaussian or RBF Kernel}
\end{table}
\begin{table}[ht!]
\centering
\begin{tabular}{l|l|l|l|l}
Dimension &     Precision &     Recall &    F measure &     Accuracy \\\hline
$40\times36$ &  0.8928  &   	0.8831 &	0.8879 &    0.8880 \\
$46\times42$ &	0.8819  &   	0.8880 &	0.8849 &    0.8840 \\
$54\times50$ &	0.8892 &    	0.8831 &	0.8861 &    0.8860 \\ \hline
\end{tabular}
\caption{\label{tab:tab3}Performance of SVM with Polynomial (Degree 3) Kernel}
\end{table}

\textbf{Comparison with Logistic Regression classifier:}\\
For comparison purpose, we also train a logistic regression classifier. Experimental results as reported in Table \ref{tab:tab3a} for this dataset suggest that the performance of the logistic regression was not in par with the SVM classifier. 

\begin{table}[ht!]
\centering
\begin{tabular}{l|l|l|l|l}
Dimension   &   Precision & Recall & F measure  & Accuracy \\\hline
$40\times36$ &  0.820   &    0.810 & 0.812      &   0.812 \\
$46\times42$ &	0.820   &    0.814 & 0.820      &   0.826 \\
$54\times50$ &	0.820   &    0.810 & 0.822      &   0.810 \\ \hline
\end{tabular}
\caption{\label{tab:tab3a}Performance of Logistic Regression classifier}
\end{table}


After comparison, we select SVM with polynomial kernel for its better performance. Once the polynomial kernel is selected, then we evaluate the performance of the polynomial kernel from degree 3 to degree 6, as shown in Table \ref{tab:tab4}. As we increase the degree of polynomial beyond degree 5, the performance degrades because of over-fitting. While in the case of a low degree polynomial, the model faces underfitting. The polynomial of degree 5 achieves an accuracy of \textbf{0.8980}.

The performance of SVM trained with the histogram of oriented (HOG) features with different cell sizes is presented in Table \ref{tab:tab5}. The HoG features are extracted from the training dataset for training SVM in four different blocks. The blocks size are $2\times2$, $4\times4$, $8\times8$, and $12\times12$. We achieve the highest accuracy of 0.8980 with a $4\times4$ block size. 

\begin{table}[ht!]
\centering
\begin{tabular}{l|l|l|l|l}
Order of Polynomial	& Precision (p)	& Recall (r)	& F-measure	& Accuracy\\ \hline
Degree 3 &	0.8928	& 0.8831	& 0.8879	& 0.8880\\ 
Degree 4 &	0.9011	& 0.8850	& 0.8930	& 0.8935\\
Degree 5 & 	0.9077	& 0.8870	& 0.8972	& 0.8980\\
Degree 6 & 	0.9079	& 0.8791	& 0.8933	& 0.8945\\
\hline
\end{tabular}
\caption{\label{tab:tab4}Polynomial Kernel Performance}
\end{table}

\begin{table}[ht!]
\centering
\begin{tabular}{l|l|l|l|l}
Cell size& Precision (p)	& Recall (r)	& F-measure	& Accuracy \\\hline
$2\times2$ &  0.8692	& 0.8692	& 0.8692	& 0.8686 \\
$4\times4$  &	0.9077	& 0.8870	& 0.8972	& 0.8980 \\
$8\times8$ &	0.8975  & 0.8761	& 0.8867	& 0.8875 \\
$12\times12$ & 0.8909	& 0.8662	& 0.8784	& 0.8795\\
\hline
\end{tabular}
\caption{\label{tab:tab5}Performance for HoG features with different cell sizes}
\end{table}

After evaluating the performance of different parameters on the proposed dataset, we select those parameters on which highest results are achieved. Those selected parameters are considered for training the classifier from training dataset. Once filter and its different parameters are selected, we then evaluate the proposed methodology on the test set, which contains 100 images of Urdu text in scene images. The text and non-text regions extracted through the channel enhanced MSER regions at text region extraction stage are fed to component analysis where two stages filtering mechanism is used to filter out non-text candidates. In first stage, we filter out non-text regions based on its geometric properties, including aspect ratio, Euler's number and stroke width variation, etc. At this stage; about 20 to 25 percent non-text region are filtered out. In second stage of filtering, SVM is trained with manually cropped text and non-text regions. In Table \ref{tab:tab6}, we compare the results of SVM trained with HOG features and unsupervised features. At component level filtering, we achieve a high f-measure score of 0.3862 of SVM trained with HOG features as shown in Table \ref{tab:tab6}.

\begin{table}[ht!]
\centering
\begin{tabular}{l|l|l|l}
Features	& Precision (p)	& Recall (r)	& F-measure \\\hline
HOG	        & 0.2609        & 	0.7432	    & 0.3862 \\
Unsupervised	& 0.2431	& 0.8021	    & 0.3731 \\
\hline
\end{tabular}
\caption{\label{tab:tab6}Performance comparison of HOG and unsupervised classification on training dataset after component filtering}
\end{table}

\begin{table}[ht!]
\centering
\begin{tabular}{l|l|l|l}
Features	& Precision (p)	& Recall (r)	& F-measure \\\hline
HOG	        & 0.5057    &	0.5473	& 0.5257 \\
Unsupervised	& 0.4938	& 0.6128	& 0.5469 \\
\hline
\end{tabular}
\caption{\label{tab:tab7}Performance comparison of HOG and unsupervised on training dataset after lines verification}
\end{table}

\begin{figure}[ht!]
\centering
\includegraphics[width=0.8\linewidth]{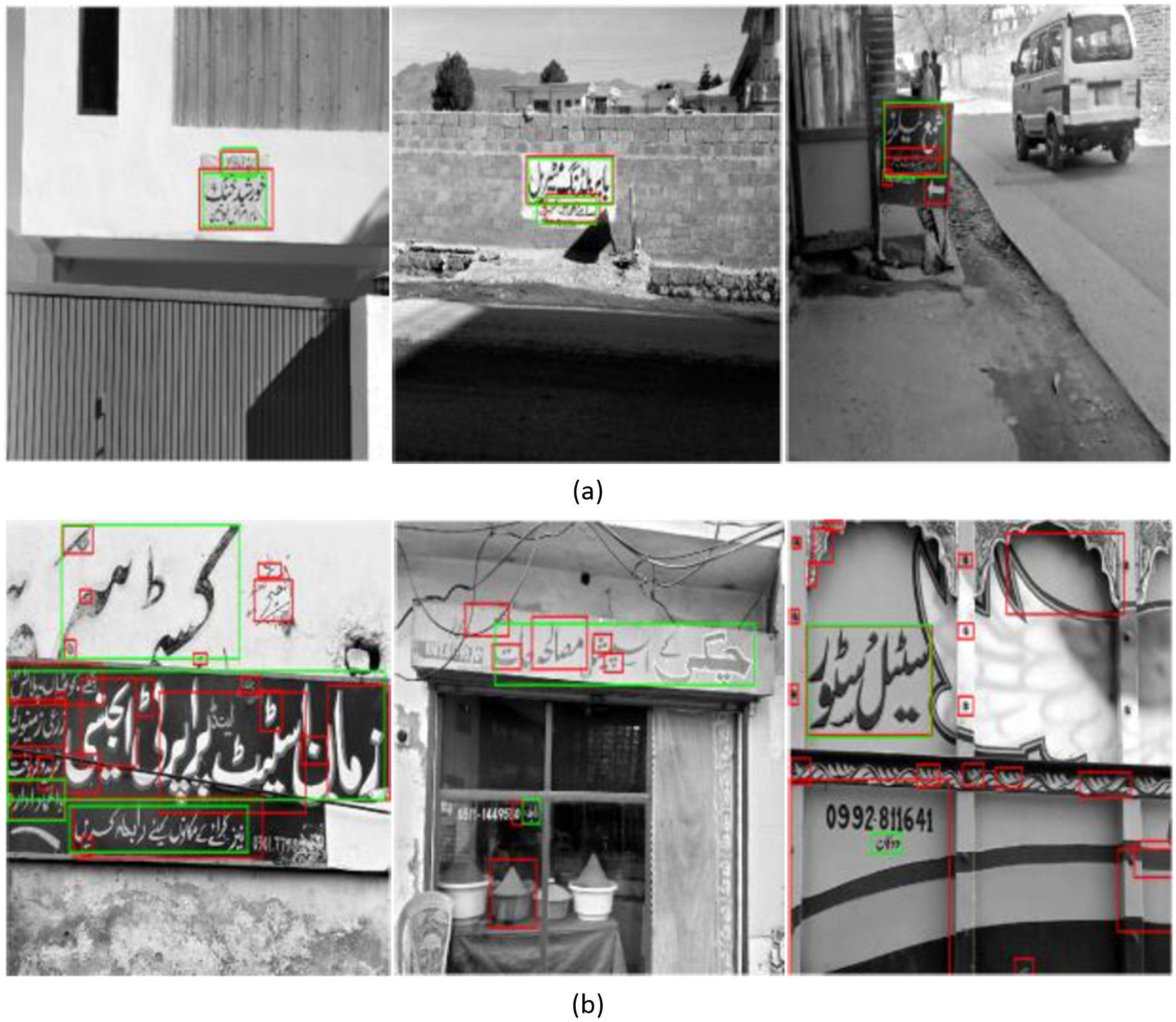}
\caption{(a) Example images for good text detection on test set images. (b) Example images for bad text detection on test set images}
\label{fig:goodbadimages}
\end{figure}

At component level filtering, most of the non-text regions along with few text regions are discarded. After component level analysis, we move on to the lines verification stage. In the lines verification stage, non-text lines are pruned by trained SVM with text and non-text lines, which are manually extracted from the training dataset. We manually extracted 700 text lines and 1400 non-text lines from the training dataset and train SVM with HOG features at that stage. The final results of the detected Urdu text lines on the proposed dataset are shown in Table \ref{tab:tab7}. We achieve a high f-measure score of 0.5469 using unsupervised features at component level filtering followed by lines verification as shown in Table \ref{tab:tab7}. Figure \ref{fig:goodbadimages} shows some samples of good and bad detection results.

\textbf{Discussion}:
Urdu text detection and recognition in natural scene images are challenging. Unlike text detection in scanned document images, the task of Urdu text detection in natural scene images has little progress. This work can provide a valuable dataset for further research work and can also serve as a baseline for making progress on this topic. The work has applications in navigation aid devices for individuals with special needs, elderly individuals and developing automatic driving technologies for roads with Urdu navigation signboards.

\section*{Conclusion and Future Directions}
In this work, we have presented a dataset for Urdu text in natural scene images. The dataset contains 500 images and is freely available for research use. This is the first dataset for Urdu text in natural scene images to the best of our knowledge. Hence, it can provide a very useful baseline for advancing the research on Urdu text extraction and recognition. Secondly, we have presented an approach for detecting Urdu text in natural scene images using MSER features extraction and an SVM classifier. For the text detection task, we have reported performance in terms of F-measure. 
Several research directions are hereby identified for the research community. The most significant is to develop a text recognition system, which may provide an end-to-end framework for Urdu contents understanding in natural scene images. Besides, the dataset size can be extended by increasing the number of images to several thousand. The dataset presented here mainly contains Urdu text in horizontal orientation. So, a significant contribution would be to add Urdu text with multiple orientations and then develop a text detection method accordingly. Since the dataset itself is now available, the performance of other techniques such as those based on neural networks and deep learning can be evaluated on the dataset. Thus a comparison can be drawn to choose the best possible model. 

\bibliography{ref3}

\begin{thebibliography}{}

\bibitem[Ahmed et~al., 2017]{ahmed2017deep}
Ahmed, S.~B., Naz, S., Razzak, M.~I., and Yousaf, R. (2017).
\newblock Deep learning based isolated arabic scene character recognition.
\newblock In {\em 2017 1st International Workshop on Arabic Script Analysis and
  Recognition (ASAR)}, pages 46--51. IEEE.

\bibitem[Ali et~al., 2020]{ali2020pioneer}
Ali, H., Ullah, A., Iqbal, T., and Khattak, S. (2020).
\newblock {Pioneer dataset and automatic recognition of Urdu handwritten
  characters using a deep autoencoder and convolutional neural network}.
\newblock {\em SN Applied Sciences}, 2(2):1--12.

\bibitem[Arafat and Iqbal, 2020]{arafat2020IEEE}
Arafat, S.~Y. and Iqbal, M.~J. (2020).
\newblock {Urdu-Text Detection and Recognition in Natural Scene Images Using
  Deep Learning}.
\newblock {\em IEEE Access}, 8:96787--96803.

\bibitem[Brooks, 2017]{brooks2017exploring}
Brooks, T.~N. (2017).
\newblock Exploring geometric property thresholds for filtering non-text
  regions in a connected component based text detection application.
\newblock {\em arXiv preprint arXiv:1709.03548}.

\bibitem[Chandio et~al., 2018]{chandio2018character}
Chandio, A.~A., Pickering, M., and Shafi, K. (2018).
\newblock Character classification and recognition for urdu texts in natural
  scene images.
\newblock In {\em 2018 International Conference on Computing, Mathematics and
  Engineering Technologies (iCoMET)}, pages 1--6. IEEE.

\bibitem[Darab and Rahmati, 2012]{darab2012hybrid}
Darab, M. and Rahmati, M. (2012).
\newblock A hybrid approach to localize farsi text in natural scene images.
\newblock {\em Procedia Computer Science}, 13:171--184.

\bibitem[Do et~al., 2021]{do2021using}
Do, D.~T., Le, T. Q.~T., and Le, N. Q.~K. (2021).
\newblock {Using deep neural networks and biological subwords to detect protein
  S-sulfenylation sites}.
\newblock {\em Briefings in Bioinformatics}, 22(3):bbaa128.

\bibitem[Epshtein et~al., 2010]{epshtein2010detecting}
Epshtein, B., Ofek, E., and Wexler, Y. (2010).
\newblock Detecting text in natural scenes with stroke width transform.
\newblock In {\em 2010 IEEE Computer Society Conference on Computer Vision and
  Pattern Recognition}, pages 2963--2970. IEEE.

\bibitem[He et~al., 2016]{he2016text}
He, T., Huang, W., Qiao, Y., and Yao, J. (2016).
\newblock Text-attentional convolutional neural network for scene text
  detection.
\newblock {\em IEEE transactions on image processing}, 25(6):2529--2541.

\bibitem[Iqbal et~al., 2017]{iqbal2017ztext}
Iqbal, K., Ali, H., Ilyas, Q.~M., Mehmood, I., Khan, H.~U., and Rehman, Z.~u.
  (2017).
\newblock Ztext: Zone based text localization in natural scene images.
\newblock {\em International Journal of Computer Science and Network Security
  (IJCSNS)}, 17(4):306.

\bibitem[Iqbal et~al., 2014]{iqbal2014bayesian}
Iqbal, K., Yin, X.-C., Hao, H.-W., Asghar, S., and Ali, H. (2014).
\newblock Bayesian network scores based text localization in scene images.
\newblock In {\em 2014 International Joint Conference on Neural Networks
  (IJCNN)}, pages 2218--2225. IEEE.

\bibitem[Iqbal et~al., 2013]{iqbal2013classifier}
Iqbal, K., Yin, X.-C., Yin, X., Ali, H., and Hao, H.-W. (2013).
\newblock Classifier comparison for mser-based text classification in scene
  images.
\newblock In {\em The 2013 International Joint Conference on Neural Networks
  (IJCNN)}, pages 1--6. IEEE.

\bibitem[Jain et~al., 2014]{jain2014text}
Jain, A., Peng, X., Zhuang, X., Natarajan, P., and Cao, H. (2014).
\newblock Text detection and recognition in natural scenes and consumer videos.
\newblock In {\em 2014 IEEE International Conference on Acoustics, Speech and
  Signal Processing (ICASSP)}, pages 1245--1249. IEEE.

\bibitem[Jamil et~al., 2011]{jamil2011edge}
Jamil, A., Siddiqi, I., Arif, F., and Raza, A. (2011).
\newblock Edge-based features for localization of artificial urdu text in video
  images.
\newblock In {\em 2011 International Conference on Document Analysis and
  Recognition}, pages 1120--1124. IEEE.

\bibitem[Khan et~al., 2012]{khan2012efficient}
Khan, K., Siddique, M., Aamir, M., and Khan, R. (2012).
\newblock {An efficient method for Urdu language text search in image based
  Urdu text}.
\newblock {\em International Journal of Computer Science Issues (IJCSI)},
  9(2):523.

\bibitem[Le et~al., 2020]{le2020xgboost}
Le, N. Q.~K., Do, D.~T., Chiu, F.-Y., Yapp, E. K.~Y., Yeh, H.-Y., and Chen,
  C.-Y. (2020).
\newblock {XGBoost improves classification of MGMT promoter methylation status
  in IDH1 wildtype glioblastoma}.
\newblock {\em Journal of Personalized Medicine}, 10(3):128.

\bibitem[Matas et~al., 2004]{matas2004robust}
Matas, J., Chum, O., Urban, M., and Pajdla, T. (2004).
\newblock Robust wide-baseline stereo from maximally stable extremal regions.
\newblock {\em Image and vision computing}, 22(10):761--767.

\bibitem[Pan et~al., 2010]{pan2010fast}
Pan, Y.-F., Liu, C.-L., and Hou, X. (2010).
\newblock Fast scene text localization by learning-based filtering and
  verification.
\newblock In {\em 2010 IEEE International Conference on Image Processing},
  pages 2269--2272. IEEE.

\bibitem[Plamondon and Srihari, 2000]{plamondon2000online}
Plamondon, R. and Srihari, S.~N. (2000).
\newblock Online and off-line handwriting recognition: a comprehensive survey.
\newblock {\em IEEE Transactions on pattern analysis and machine intelligence},
  22(1):63--84.

\bibitem[Raza and Siddiqi, 2012]{raza2012database}
Raza, A. and Siddiqi, I. (2012).
\newblock A database of artificial urdu text in video images with
  semi-automatic text line labeling scheme.
\newblock In {\em Proc. 4th Int. Conf. Adv. Multimedia (MMEDIA)}, pages 75--81.

\bibitem[Shahab et~al., 2011]{shahab2011icdar}
Shahab, A., Shafait, F., and Dengel, A. (2011).
\newblock Icdar 2011 robust reading competition challenge 2: Reading text in
  scene images.
\newblock In {\em 2011 international conference on document analysis and
  recognition}, pages 1491--1496. IEEE.

\bibitem[Shi et~al., 2013]{shi2013scene}
Shi, C., Wang, C., Xiao, B., Zhang, Y., and Gao, S. (2013).
\newblock Scene text detection using graph model built upon maximally stable
  extremal regions.
\newblock {\em Pattern recognition letters}, 34(2):107--116.

\bibitem[Yan et~al., 2017]{yan2017effective}
Yan, C., Xie, H., Liu, S., Yin, J., Zhang, Y., and Dai, Q. (2017).
\newblock {Effective Uyghur language text detection in complex background
  images for traffic prompt identification}.
\newblock {\em IEEE transactions on intelligent transportation systems},
  19(1):220--229.

\bibitem[Yao et~al., 2012]{yao2012detecting}
Yao, C., Bai, X., Liu, W., Ma, Y., and Tu, Z. (2012).
\newblock Detecting texts of arbitrary orientations in natural images.
\newblock In {\em 2012 IEEE conference on computer vision and pattern
  recognition}, pages 1083--1090. IEEE.

\bibitem[Yin et~al., 2012]{yin2012effective}
Yin, X., Yin, X.-C., Hao, H.-W., and Iqbal, K. (2012).
\newblock Effective text localization in natural scene images with mser,
  geometry-based grouping and adaboost.
\newblock In {\em Proceedings of the 21st International Conference on Pattern
  Recognition (ICPR2012)}, pages 725--728. IEEE.

\bibitem[Yuan et~al., 2016]{yuan2016incremental}
Yuan, Y., Xiong, Z., and Wang, Q. (2016).
\newblock An incremental framework for video-based traffic sign detection,
  tracking, and recognition.
\newblock {\em IEEE Transactions on Intelligent Transportation Systems},
  18(7):1918--1929.

\bibitem[Zhang et~al., 2013]{zhang2013text}
Zhang, H., Zhao, K., Song, Y.-Z., and Guo, J. (2013).
\newblock Text extraction from natural scene image: A survey.
\newblock {\em Neurocomputing}, 122:310--323.

\bibitem[Zheng et~al., 2010]{zheng2010text}
Zheng, Q., Chen, K., Zhou, Y., Gu, C., and Guan, H. (2010).
\newblock Text localization and recognition in complex scenes using local
  features.
\newblock In {\em Asian Conference on Computer Vision}, pages 121--132.
  Springer.

\bibitem[Zhu et~al., 2016]{zhu2016scene}
Zhu, Y., Yao, C., and Bai, X. (2016).
\newblock Scene text detection and recognition: Recent advances and future
  trends.
\newblock {\em Frontiers of Computer Science}, 10(1):19--36.

\end{thebibliography}

\end{document}